\documentclass[a4paper, 10pt, conference]{ieeeconf}
\IEEEoverridecommandlockouts

\usepackage{cite}
\usepackage{amsmath,amssymb,amsfonts}
\usepackage{algorithmic}
\usepackage{graphicx}
\usepackage{textcomp}
\usepackage{xcolor}

\title{\LARGE \bf
Robotic Multimodal Data Acquisition for In-Field Deep Learning Estimation of Cover Crop Biomass
}

\author{Joe Johnson$^{1}$, Phanender Chalasani$^{2}$, Arnav Shah$^{3}$, Ram L. Ray$^{4}$, and Muthukumar Bagavathiannan$^{1}$
\thanks{$^{1}$ Joe is a PhD student in the Department of Soil and Crop Sciences, Texas A\&M University, College Station, TX 77840, USA
        {\tt\small joejohnson2905@tamu.edu}}%
\thanks{$^{2}$ Phanender is a MS student in the Department of Computer Science and Engineering, Texas A\&M University, College Station, TX 77840, USA
        {\tt\small phanenderchalasani@tamu.edu}}%
\thanks{$^{3}$ Arnav is an undergraduate student in the Department of Mechanical Engineering, Texas A\&M University, College Station, TX 77840, USA
        {\tt\small arnav.s2@tamu.edu}}%
\thanks{$^{4}$ Ram is a Professor in the College of Agriculture, Food and Natural Resources, Prairie View A\&M University, Prairie View, TX 77446, USA
        {\tt\small raray@pvamu.edu}}%
\thanks{$^{1}$ Muthukumar is a Professor in the Department of Soil and Crop Sciences, Texas A\&M University, College Station, TX 77840, USA
        {\tt\small m.bagavathiannan@ag.tamu.edu}}%
}

\begin{document}

\maketitle
\thispagestyle{empty}
\pagestyle{empty}

\begin{abstract}
Accurate weed management is essential for mitigating significant crop yield losses, necessitating effective weed suppression strategies in agricultural systems. Integrating cover crops (CC) offers multiple benefits, including soil erosion reduction, weed suppression, decreased nitrogen requirements, and enhanced carbon sequestration, all of which are closely tied to the aboveground biomass (AGB) they produce. However, biomass production varies significantly due to microsite variability, making accurate estimation and mapping essential for identifying zones of poor weed suppression and optimizing targeted management strategies. To address this challenge, developing a comprehensive CC map, including its AGB distribution, will enable informed decision-making regarding weed control methods and optimal application rates. Manual visual inspection is impractical and labor-intensive, especially given the extensive field size and the wide diversity and variation of weed species and sizes. In this context, optical imagery and Light Detection and Ranging (LiDAR) data are two prominent sources with unique characteristics that enhance AGB estimation. This study introduces a ground robot-mounted multimodal sensor system designed for agricultural field mapping. The system integrates optical and LiDAR data, leveraging machine learning (ML) methods for data fusion to improve biomass predictions. The best ML-based model for dry AGB estimation achieved an coefficient of determination ($R^2$) value of 0.88, demonstrating robust performance in diverse field conditions. This approach offers valuable insights for site-specific management, enabling precise weed suppression strategies and promoting sustainable farming practices. The integration of high-resolution optical and LiDAR data from a robot-mounted system, combined with ML techniques, establishes a scalable framework for automated biomass estimation in large-scale agricultural field conditions, enhancing decision-making in precision agriculture.

\end{abstract}

\section{Introduction}
The excessive use of herbicides in no-till and reduced tillage systems, coupled with a lack of new herbicide chemistries, has led to widespread weed resistance \cite{b8}. To balance weed control with soil conservation, integrated management strategies such as winter CCs are increasingly vital \cite{b10}.

Cover crop biomass plays a crucial role in weed suppression, as high biomass levels physically inhibit weeds by competing for light, space, water, and nutrients while blocking seedling establishment \cite{b11}. In cereal rye (Secale cereale L.), a widely used winter CC in the USA, the recommended dry aboveground biomass for effective weed control is approximately 0.8 Kg $m^{-2}$, though lower levels is approximately 0.3 Kg $m^{-2}$) have also shown efficacy \cite{b12}. However, biomass production is highly variable, influenced by climate, soil fertility, and management practices \cite{b13}.

To optimize weed suppression and support management decisions, growers need efficient and cost-effective CC biomass estimation methods \cite{b14}. Traditional biomass measurements involve destructive sampling and significant labor, while indirect indices such as leaf area index (LAI) and photosynthetically active radiation (PAR) require canopy-level sampling, making them impractical for large-scale assessments \cite{b15}.

\begin{figure}[h!]
\centering
\includegraphics[height=0.23\textheight]{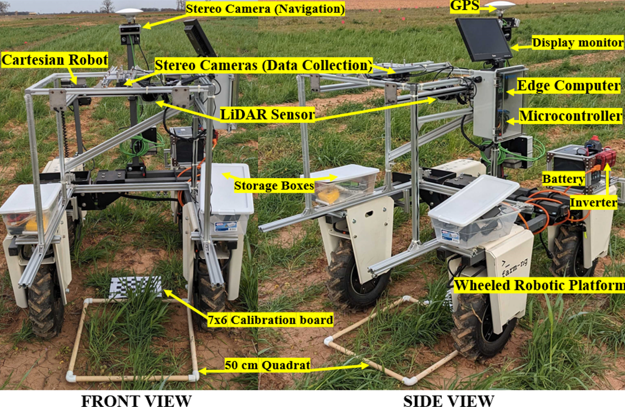}
\caption{Setup of a robotic multimodal data collection system over a unmanned ground vehicle and main components.} \label{fig1}
\end{figure}

Accurate estimation of CC AGB is critical for predicting and optimizing weed suppression \cite{b1}. Combining plant height maps with RGB image data improves biomass predictions using data-driven methods \cite{b2}. Autonomous robotic platforms present an ideal solution for continuous, nondestructive field data collection \cite{b3}.

This research focuses on developing, calibrating, and testing automated hardware and software systems for CC AGB estimation in field conditions, offering insights into its practicality and scalability \cite{b5}. A modular ground robotic platform is designed for efficient multimodal data collection, as illustrated in Fig. \ref{fig1}. Integrating stereo-based depth estimation with RGB data enhances ML models for dry biomass prediction. The study aims to develop a robotic system for field-based multimodal data collection to acquire RGB-depth information and to train and test ML models utilizing multimodal inputs for precise CC dry AGB estimation. 

\section{Materials and Methods}

\subsection{Experimental setup}

\begin{figure}[h!]
\centering
\includegraphics[height=0.2\textheight]{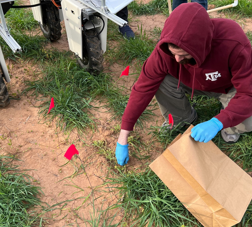}
\caption{The scene describing the manual destructive biomass sampling process of cutting the aboveground biomass within the quadrat area and bagging.} \label{fig2}
\end{figure}

Data for this study were collected from the Texas A\&M AgriLife Research farm near College Station, TX (30.537618 N, 96.427028 W; elevation: 70m) in an area of 4,000 square meters. The CC species cereal rye (Secale cereale L.) is grown at this specific farm location. A Randomized complete block design with a factor varying CC planting density (0.25X, 0.75X, and 1.5X; were the planting density X = 75 pounds per acre) and three replications per planting density with each plot size $20 m \times 12 m$. A total of 27 plots and 135 samples were collected. The AGB data were collected by destructive sampling of CCs within the rectangular quadrat ($0.5 m \times 0.5 m$) in spring 2024 on different dates. The dates of sampling of these CC are 02/28/2024, 03/03/2024 and 04/02/2024. The biomass sampling process involved manually cutting the aboveground biomass within the quadrat area, as illustrated in Fig. \ref{fig2}, followed by bagging it in paper collection bags. The biomass was then oven-dried before being weighed to obtain the dry AGB ground truth.

\subsection{Hardware System and data collection}

The design and development of a 2-DoF Cartesian robotic platform, capable of supporting a maximum payload of 1 kg, is integrated with the Farm-NG Amiga wheeled robot (450 kg payload). This platform is equipped with two Luxonis OAK-D RGB-depth sensors (12 MP RGB and 1 MP stereo pair resolution) and an RPLIDAR S2 LiDAR (0.1° angular resolution and 12 m range) to capture high-quality multimodal data in field conditions. The 2-DoF Cartesian robotic platform facilitates 2D LiDAR scanning, both while stationary and in motion, ensuring precise environmental mapping regardless of the movement of the ground robot.

\begin{figure}[h]
\centering
\includegraphics[height=0.28\textheight]{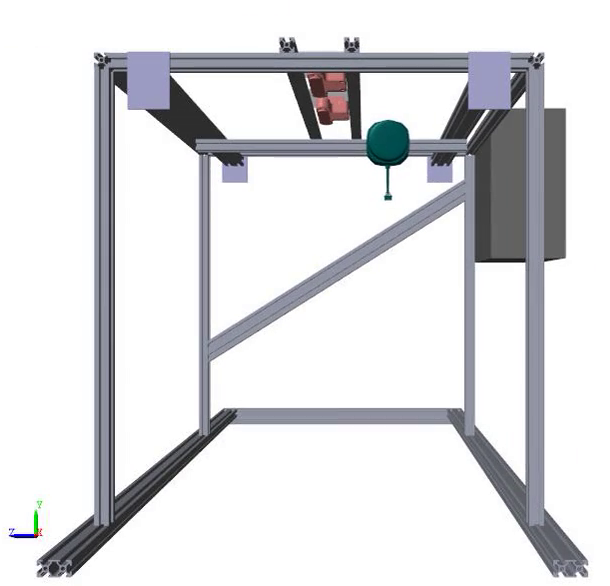}
\caption{The CAD model of the 2-DoF Cartesian robotic system with multimodal sensors used for data collection.} \label{fig3}
\end{figure}

The Cartesian robotic platform, with its CAD model shown in Fig. \ref{fig3}, is designed to capture RGB-depth and LiDAR data from a height of 1.5 meters while meeting the minimum data acquisition requirements of its sensors. Specifically, the stereo camera requires at least 0.85 m of clearance from the target (e.g., plant canopy) to ensure accurate depth perception. Hence, the maximum target plant height can be 0.65 m which is ideal for popular CC species. The LiDAR is mounted perpendicular to the field surface on the Cartesian robotic platform to optimize scanning performance, positioned at a height of 1.45 meters, aligned perpendicular to the ground, and configured to operate at their maximum resolution.

\begin{figure}[h]
\centering
\includegraphics[height=0.195\textheight]{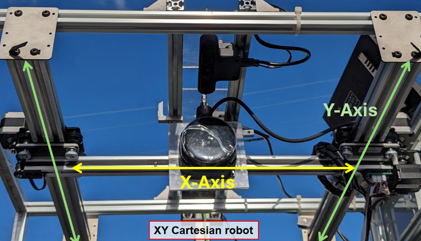}
\caption{A close-up view of the 2-DoF Cartesian robotic system with multimodal sensors attached.} \label{fig4}
\end{figure}

The multimodal sensor mounting platform is designed with dimensions of 0.90 m in height, 1.10 m in width, and 0.85 m in depth. A close-up view of the sensors on the Cartesian robotic platform is provided in Fig. \ref{fig4}. The frame is constructed using 20 mm × 20 mm T-slotted aluminum extrusion rods and 5 mm acrylic sheets, ensuring secure sensor placement. The aluminum frame is selected for its lightweight, durability, and ease of fabrication, making it well-suited for prototyping applications. 

\begin{figure}[h]
\centering
\includegraphics[height=0.3\textheight]{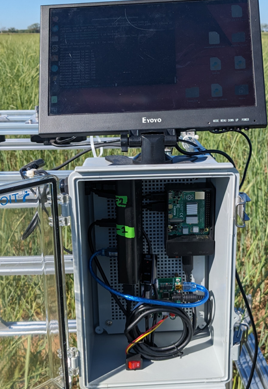}
\caption{The enclosure for the auxiliary single-board computer (Raspberry Pi 4B), USB hub, and Cartesian robot drive to protect from the adverse field conditions.} \label{fig5}
\end{figure}

An electrical box of IP67 grade was used to house the auxiliary single-board computer (Raspberry Pi 4B), USB hub, and Cartesian robot drive, as shown in Fig. \ref{fig5}. This box protects sensitive components from dust and humidity of agricultural fields. 

\begin{figure}[h]
\centering
\includegraphics[height=0.2\textheight]{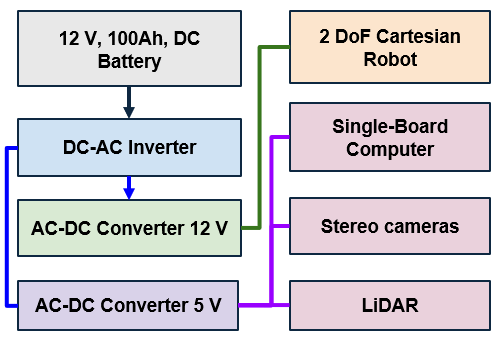}
\caption{The block diagram of the electrical power distribution for the robotic data collection system platform.} \label{fig6}
\end{figure}

The sensor platform mounted on the Amiga robot features a modular design with an integrated auxiliary power source. This adaptable setup allows for easy deployment and seamless configuration across various mobile robotic platforms. A high-level block diagram illustrating the electrical power distribution is shown in Fig. \ref{fig6}.

For automatic multimodal data collection, the ground robot operates at a linear speed of 34 ft/min (0.173 m/s) while navigating crop rows to ensure high-quality data acquisition. This speed facilitates more than 95\% overlap at a capture rate of five frames per second, minimizing motion artifacts and preventing jerks caused by uneven agricultural terrain. RGB video footage, along with recordings from two monochrome sensors, is collected for 10 seconds at 15 frames per second using two OAK-D RGB-depth cameras. In total, 24,570 paired RGB and depth data samples were used for training and testing ML models.

\begin{figure}[h]
\centering
\includegraphics[height=0.24\textheight]{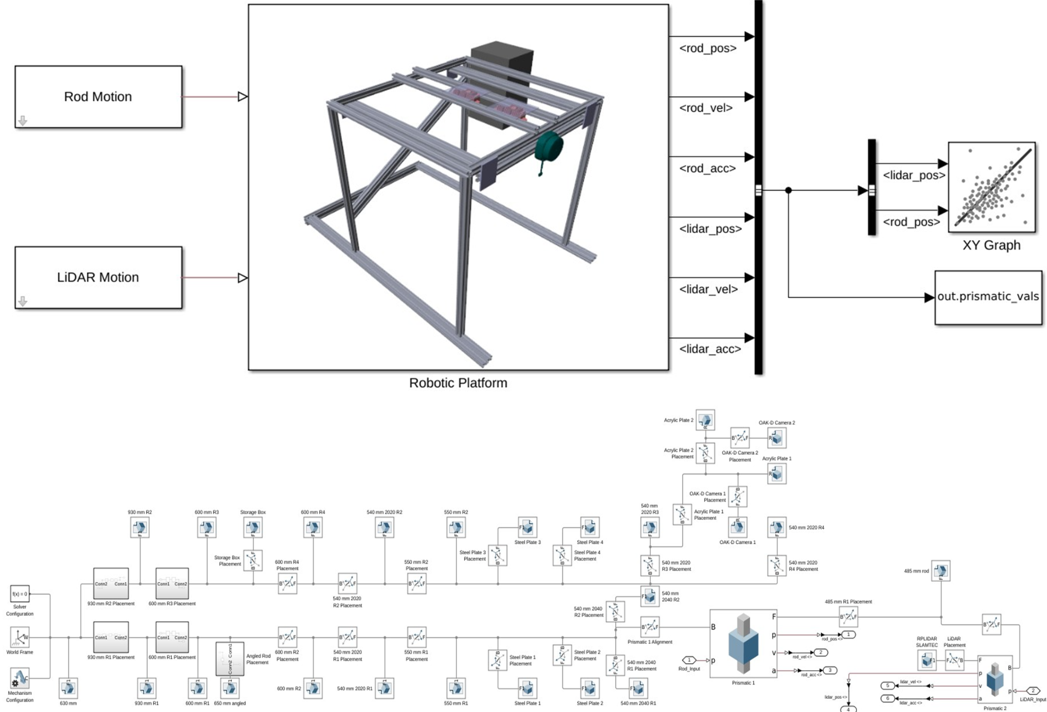}
\caption{The block diagram of the MATLAB Simulink-based for multibody simulation of the robotic sensor platform.} \label{fig7}
\end{figure}

A MATLAB-based Simscape Multibody model is used to simulate a two-DoF Cartesian mechanical system for LiDAR data collection in static robot conditions. This simulation enables the evaluation of the sensor’s working volume and assists in planning the LiDAR scanning path and motion steps to optimize data collection efficiency. The block diagram of the MATLAB-based Simscape Multibody simulation is shown in Fig. \ref{fig7}.

\subsection{Video Processing and Data Analysis}
Using Python (Python Software Foundation, Wilmington, DE, USA), we processed each recorded video (15 FPS) by extracting 70 selected frames to generate RGB images and its corresponding dense depth maps from disparity maps using stereo vision. Depth maps corresponding to PVC marker locations were identified for analysis. The processing workflow included the following steps: determining video type (RGB or disparity map video), accessing the video path, reading the video file, retrieving the total frame count, extracting all frames, checking whether the current frame matches a predefined list, and finally converting the selected frames into the desired image formats.

\begin{figure}[h]
\centering
\includegraphics[height=0.14\textheight]{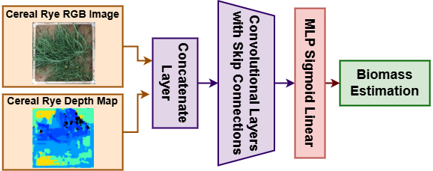}
\caption{A High-level block diagram of the multimodal data fusion based ResNet-50 DL model.} \label{fig8}
\end{figure}

The AGB of CC was analyzed using four ML models: Random Forest Regression (RFR), Support Vector Regression (SVR) \cite{b6}, Artificial Neural Network (ANN), and a custom deep learning (DL) model. The custom DL architecture employs ResNet50 as its backbone with a sigmoid linear output layer, as illustrated in Fig. \ref{fig8}.

The CC dry AGB models was validated for biomass prediction using independently collected field data. Predicted biomass values were generated by applying the four ML models, which leverage correlations between color and depth information with the measured biomass for each 0.25 $m^2$ quadrat.

In the preprocessing step for data analysis, the RGB and depth map are downsampled to 512 x 512 pixels. All architectures were implemented in Python language using the TensorFlow 2.10.1 framework and Keras API 2.10.0 with similar hyperparameters. The CuDNN (8.0.1.77) library was installed to best utilize the graphics card performance. The hyperparameters are the following, a batch size of 16, a learning rate of 0.001, and training for 15 epochs. All ML models were trained and tested on a desktop workstation with an Intel(R) Core (TM) i9- 10900X CPU with a speed of 3.70GHz, 64GB of RAM, 1 TB storage drive and enhanced with dual graphics card (NVIDIA RTX A4000) with 16 GB memory. 

\subsection{Model Evaluation}
The performance and reliability of linear regression models, developed by fitting extracted phenotypic traits both individually and in combination, were assessed using the coefficient of determination ($R^2$), root mean square error (RMSE), and relative root mean square error (RRMSE). These metrics are defined as follows:

\begin{equation}
R^2 = 1 - \frac{\sum_{i=1}^{n} (y_i - \hat{y}_i)^2}{\sum_{i=1}^{n} (y_i - \bar{y})^2}
\end{equation}

\begin{equation}
RMSE = \sqrt{\frac{1}{n} \sum_{i=1}^{n} (y_i - \hat{y}_i)^2}
\end{equation}

\begin{equation}
RRMSE = \frac{RMSE}{\bar{y}} \times 100
\end{equation}

where $y_i$ is the actual value, $\hat{y}_i$ is the predicted value, $n$ is the number of observations, $\bar{y}$ is the mean of actual values, providing a relative measure of error in percentage form.

\section{RESULTS AND DISCUSSION}

The stereo vision-based depth map achieved a resolution of 1 mm at an imaging height of 1.5 m. However, as the object’s distance from the stereo cameras increased, the predicted height resolution progressively declined. Additionally, depth information generation was influenced by the stereo camera’s exposure settings \cite{b7}. Accurate biomass estimation was achieved by integrating color, texture, shape, and plant structure data. Among the trained models, the ResNet-50 based biomass regression model demonstrated the best performance, as shown in Table \ref{table4}.

The $R^2$ for all tested models ranged from 0.65 to 0.88. In the initial biomass sampling, regression exhibited a positive linear trend up to 0.25 kg/m², beyond which CC pixel density remained unchanged despite continued biomass accumulation. Because biomass accumulation showed a continuous positive correlation with crop height within the observed range, we proposed that combining CC pixel density and crop height would improve biomass prediction accuracy. Our results indicate that effective modeling for this technology must account for the unique morphological characteristics of each CC species. Specifically, species with high light interception in those with horizontally arranged leaves may have lower predictive accuracy compared to grasses with vertically oriented foliage.

The effectiveness of sensing technology is closely linked to the volume of collected data, with accuracy and speed forming an inherent tradeoff. Striking the right balance is essential for optimizing data acquisition methods and minimizing potential errors. While a larger dataset enhances predictive accuracy, it also leads to increased processing time. One key challenge is image collection speed. Faster data capture expands coverage but also introduces vibrations and fluctuations in the camera’s height relative to the canopy. If left unaddressed, these issues may become more pronounced when adapting the technology for large-scale agricultural applications, underscoring the need for effective corrective strategies. 

\begin{table}[h]
\caption{The comparison of ML-based dry AGB estimation using multimodal data}
\label{table4}
\renewcommand{\arraystretch}{1.5}
\begin{center}
\begin{tabular}{|l||c|c|c|}
\hline
\textbf{ML Model} & \textbf{R}$^2$ & \textbf{RMSE} (kg/$m^2$) & \textbf{RRMSE} (\%)\\
\hline
RFR  & 0.65 & 0.0433 & 14.82\\
\hline
SVR  & 0.76 & 0.0372 & 12.74\\
\hline
ANN & 0.81 & 0.0301 & 10.31 \\
\hline
ResNet 50 & 0.88 & 0.0289 & 9.91\\
\hline
\end{tabular}
\end{center}
\end{table}

\section{CONCLUSIONS AND FUTURE WORK}
Accurately measuring CC biomass is essential for predicting weed suppression efficacy, and other ecosystem benefits. However, conventional biomass measurement methods are often labor-intensive, costly, and limited in their ability to capture in-field variability. This study presents the development of a Cartesian robotic platform designed to enable precise multimodal sensor data collection for agricultural applications. While still in its development phase, the modular platform demonstrates significant utility by capturing high-quality multimodal sensor data in both static and dynamic field conditions. Its modular design, constructed from readily available aluminum components and a cost-effective off-the-shelf module, enhances adaptability and scalability. Furthermore, the hardware requires no welding, allowing for easy replacement or reconfiguration of frame components to suit different research and field conditions. 

The findings highlight the effectiveness of ML models that integrate both RGB and depth data, consistently outperforming single-modality approaches in biomass estimation. The ResNet 50 based DL AGB estimation model achieved high predictive accuracy, underscoring the potential of multimodal DL for agricultural field analysis.

Future advancements should focus on improving localization and size estimation of plant organs using 3D DL models, enabling precise herbicide application based on accurate 3D measurements. To enhance model robustness and adaptability, validation across diverse sites and CC growth stages is essential. Additionally, developing smartphone-based applications for real-time data collection and decision-making can empower farmers and advisors with accessible precision agriculture tools. Investigating spatial variability in biomass distribution within and across fields will further refine precision management strategies. Integrating imagery from unmanned aerial systems and satellite data for large-scale biomass estimation will contribute to a scalable framework for automated assessments, ultimately enhancing data-driven decision-making in precision agriculture.

Overall, this study demonstrates that combining crop structural information with spectral data enables more accurate crop biomass estimation compared to other non-destructive methods, reinforcing the value of multimodal sensing and ML for agricultural applications. 

\section*{}
{\small
\section*{ACKNOWLEDGEMENT}
This project was supported by the Panther Research \& Innovation for Scholarly Excellence (PRISE) grant program (2023–2024) of the Texas A\&M University (TAMU) System, for which the authors express their sincere gratitude. Special thanks to the TAMU Weed Science team for their invaluable assistance throughout this project. The authors also acknowledge the resources and support provided by the Aggie Research Mentoring Program (ARP) at Texas A\&M University.

\vspace{12pt}


\end{document}